\pdfoutput=1
\PassOptionsToPackage{table,xcdraw}{xcolor}
\documentclass{article}
\usepackage{iclr2027_conference,times}
\iclrfinalcopy

\usepackage[utf8]{inputenc}
\usepackage[T1]{fontenc}
\usepackage{hyperref}
\usepackage{url}
\usepackage{booktabs}
\usepackage{amsfonts}
\usepackage{nicefrac}
\usepackage{microtype}
\usepackage{xcolor}

\usepackage{graphicx}
\usepackage{subcaption}
\usepackage{caption}
\usepackage{amsmath}
\usepackage{amssymb}
\usepackage{mathtools}
\usepackage{amsthm}
\usepackage{enumitem}
\usepackage[capitalize,noabbrev]{cleveref}
\usepackage{placeins}
\usepackage{multirow}
\usepackage{pifont}
\usepackage{algorithm}
\usepackage{algorithmic}
\usepackage{wrapfig}
\newcommand{\imp}[2]{#1\hspace{5pt}$\uparrow$\hspace{1pt}{\scriptsize #2}}
\newcommand{\cmark}{\ding{51}}
\newcommand{\xmark}{\ding{55}}
\renewcommand\thesubfigure{(\alph{subfigure})}

\newcommand{\algo}{MARS}
\newcommand{\res}[2]{#1\textsubscript{#2}}

\usepackage{needspace}
\makeatletter
\let\mars@wrapfit\WF@putfigmaybe
\def\WF@putfigmaybe{\ifinner\else\Needspace{\WF@size}\fi\mars@wrapfit}
\makeatother

\title{Unsupervised Long-Tailed Adaptation of \\ Vision-Language Models}

\author{%
  Keliang Chen$^{1}$\thanks{Equal contribution.} \quad
  Yaxin Hou$^{1}$\footnotemark[1] \quad
  Hui Liu$^{3}$ \quad
  Yuheng Jia$^{1,2,3}$\thanks{Corresponding author.} \\
  $^{1}$School of Computer Science and Engineering, Southeast University, Nanjing 210096, China \\
  $^{2}$Key Laboratory of New Generation Artificial Intelligence Technology and Its \\
  \phantom{$^{2}$}Interdisciplinary Applications (Southeast University), Ministry of Education, China \\
  $^{3}$School of Computing and Information Sciences, Saint Francis University, Hong Kong, China \\
  \texttt{\{220255105,yaxin,yhjia\}@seu.edu.cn}, \ \texttt{h2liu@sfu.edu.hk}
}

\begin{document}

\maketitle
\lhead{}
\renewcommand{\headrulewidth}{0pt}

\begin{abstract}

Adapting vision-language models to downstream tasks has achieved remarkable success by leveraging pseudo-labels generated from unlabeled data. Existing methods typically assume a uniform unlabeled data distribution, and thus the resulting pseudo-label distribution is likewise uniform. However, real-world data distributions are often long-tailed. To tackle this, we formalize a new scenario termed Unsupervised Long-Tailed Adaptation (ULTA). Under this scenario, existing methods exhibit a contrasting phenomenon: head-class performance drops sharply, which is distinct from supervised long-tailed learning where tail classes suffer the most. In particular, we uncover that the distributional mismatch not only erodes head-class boundaries, but also pushes head samples into confusable classes, reinforcing the model's inherent bias. To address these issues, we propose a novel model called Margin-Aware Refinement with Structural alignment (MARS). Specifically, we mitigate head-class boundary erosion via Boundary-Preserving Alignment, which takes the zero-shot VLM as a fixed visual reference to suppress probability increases that lack visual support in the training targets. Building upon this, we introduce Margin-aware Self-Refinement, which employs a dynamic adjustment strategy to refine tail and confusable classes while preventing prediction bias. Extensive experiments on nine benchmark datasets demonstrate that MARS outperforms state-of-the-art methods, achieving an average accuracy improvement of \textbf{4.71} percentage points.

\end{abstract}

\section{Introduction}
\label{sec:intro}

Vision-language models (VLMs)~\citep{jia2021align, li2022blip, xiao2024florence2}, such as CLIP~\citep{radford2021clip}, have demonstrated remarkable zero-shot capabilities by establishing a shared feature space across visual and textual modalities via web-scale data.
Despite the impressive zero-shot capabilities, their performance is limited when transferred to specific downstream tasks due to the domain gap. While fine-tuning on labeled data can bridge this gap~\citep{zhou2022coop, jia2022vpt, ouali2023lfa, shi2024candle}, acquiring sufficient annotations in specialized scenarios (e.g., medical imaging~\citealp{huang2021gloria, wang2022mgca}) is impractical. Consequently, recent studies~\citep{huang2022upt, mirza2023lafter, menghini2023grip} have shifted their focus toward enhancing performance by exploiting unlabeled data.

\begin{figure}[!t]
    \centering
    {
    \captionsetup[subfigure]{
        labelformat=simple,
        labelfont={normalfont},
        textfont={normalfont},
        skip=4pt,
        justification=centering
    }

    \begin{subfigure}[t]{0.25\columnwidth}
        \centering
        \includegraphics[width=\linewidth]{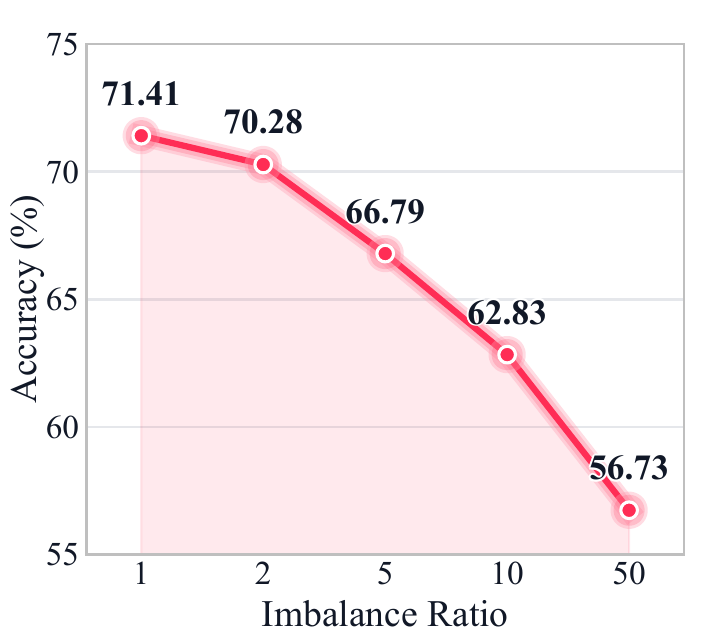}
        \caption{}
        \label{fig:1_a}
    \end{subfigure}
    \hspace{-0.1cm}%
    \begin{subfigure}[t]{0.25\columnwidth}
        \centering
        \includegraphics[width=\linewidth]{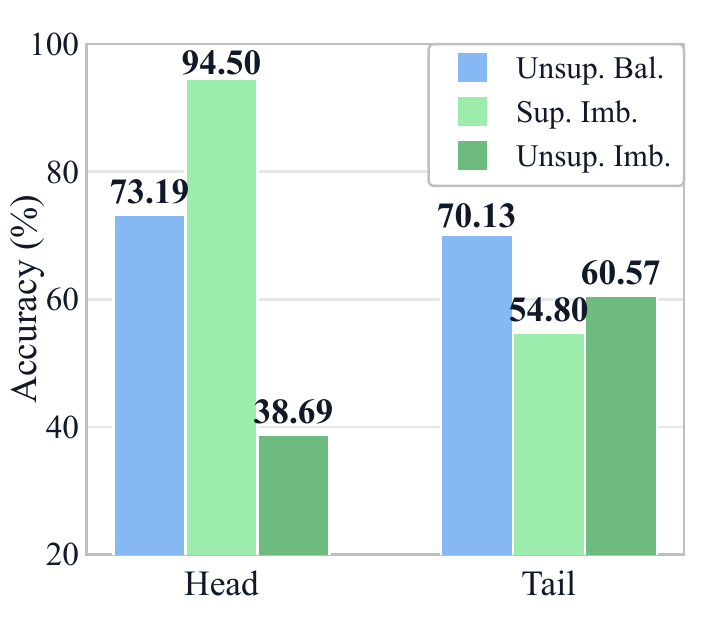}
        \caption{}
        \label{fig:1_b}
    \end{subfigure}
    \hspace{-0.1cm}%
    \begin{subfigure}[t]{0.25\columnwidth}
        \centering
        \includegraphics[width=\linewidth]{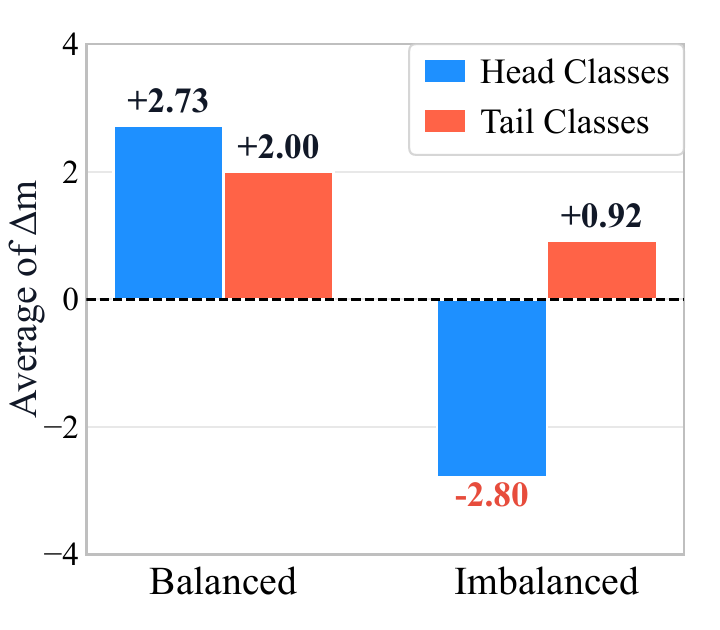}
        \caption{}
        \label{fig:2_a}
    \end{subfigure}
     \hspace{-0.1cm}%
    \begin{subfigure}[t]{0.25\columnwidth}
        \centering
        \includegraphics[width=\linewidth]{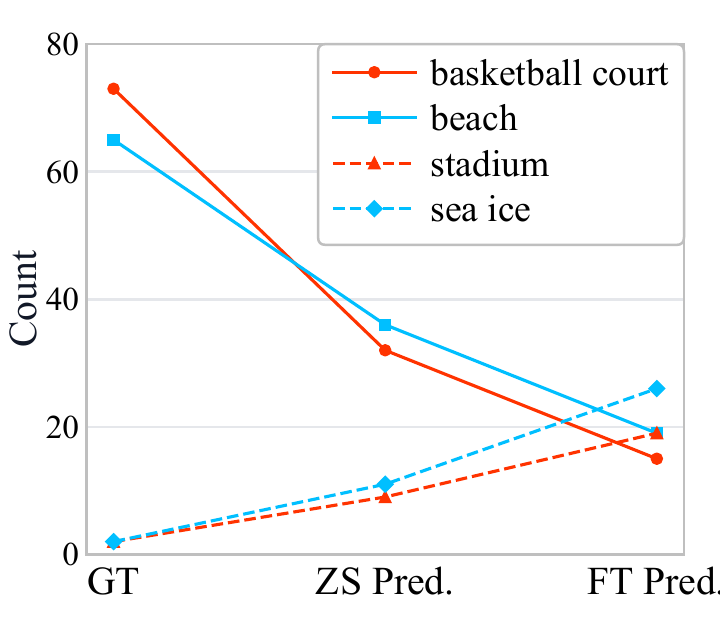}
        \caption{}
        \label{fig:2_b}
    \end{subfigure}
    }

    \vspace{-0.2cm}

    \caption{
    \textbf{(a)} The overall test accuracy of the existing unsupervised method CPL~\citep{zhang2024cpl} under different imbalance ratios.
    \textbf{(b)} Comparison of head- and tail-class accuracies across three settings on the RESISC45 dataset: unsupervised learning with balanced data (Unsup. Bal.), supervised learning with imbalanced data (Sup. Imb.), and unsupervised learning with imbalanced data (Unsup. Imb.).
    \textbf{(c)} Comparison of average logit margin changes ($\Delta m = \text{margin}_{\text{Fine-tuned}} - \text{margin}_{\text{Zero-shot CLIP}}$). Under the long-tailed setting, the head-class margin significantly decreases compared to the uniform setting, indicating severe boundary erosion.
    \textbf{(d)} Label counts across three cases: ground-truth (GT), zero-shot predictions (ZS Pred.), and fine-tuned predictions (FT Pred.).
    We group confusable pairs by color (e.g., blue: ``beach'' vs. ``sea ice'') and observe that zero-shot CLIP inherently exhibits bias toward confusable classes, which is further exacerbated by fine-tuning.
    }
    \label{fig:1}
    \label{fig:2}

    \vspace{-0.2cm}
\end{figure}

These studies~\citep{zhang2024cpl, li2025tmp, wang2025cap} primarily derive supervisory signals from unlabeled data via pseudo-labeling. Those methods usually leverage a uniform prior explicitly or implicitly to yield class-balanced pseudo-labels. However, real-world data exhibits a long-tailed distribution, creating a mismatch with this uniform prior. As evidenced in Fig.~\ref{fig:1_a}, these methods (e.g., CPL~\citealp{zhang2024cpl}) suffer from substantial performance degradation on long-tailed data. Notably, we observe an inverted degradation pattern that diverges from supervised long-tailed learning. In supervised long-tailed learning, the tail classes typically suffer the most due to the scarcity of their training samples. While in unsupervised long-tailed scenarios, existing methods lead to \textbf{a severe performance drop for head classes} (see Fig.~\ref{fig:1_b}).

Empirically, we find that in long-tailed scenarios, the uniform assumption in existing methods forces the model to misclassify many head samples as tail classes. Fitting these ``head-as-tail'' false samples leads to head-class decision boundary erosion, which causes a significant decrease in the logit margin (i.e., the gap between the ground-truth logit and the highest non-target logit) of head classes (see Fig.~\ref{fig:2_a}). Furthermore, zero-shot CLIP's inherent bias toward confusable classes (e.g., semantically similar pairs like ``beach'' vs. ``sea ice'') is exacerbated during fine-tuning (see Fig.~\ref{fig:2_b}).

To address these challenges, we propose the Margin-Aware Refinement with Structural alignment (MARS) to tackle \textbf{Unsupervised Long-Tailed Adaptation (ULTA)}, a realistic scenario where VLMs are adapted to unlabeled data with a long-tailed distribution. MARS follows a preserve-before-refine design: it first preserves the reliable head-class boundary and then refines the tail and confusable classes.
Specifically, to mitigate head-class boundary erosion caused by the erroneous assignment of pseudo-labels, we propose Boundary-Preserving Alignment (BPA). This approach exploits the class-level visual structure of zero-shot CLIP as a fixed reference and suppresses probability increases unsupported by visual evidence before they become training targets, instead of directly fitting the model's own erroneous predictions. Building upon this, we introduce Margin-aware Self-Refinement (MSR) to further improve tail and confusable classes while preserving the well-learned decision boundary. Adopting the aligned model as a teacher, this strategy leverages the estimated class distribution and the sample-specific margin to dynamically adjust the learning focus, distinguishing valid supervision from noise to actively acquire discriminative features.

In summary, our contributions are as follows:
\begin{itemize}[leftmargin=12pt]

    \item To our knowledge, we are the first to study unsupervised long-tailed adaptation (ULTA), a realistic yet under-explored scenario where VLMs are adapted to long-tailed unlabeled data.

    \item We uncover a distinct phenomenon where existing unsupervised methods suffer severe degradation on head classes due to boundary erosion from forced uniform pseudo-labels, and observe aggravated prediction bias in confusable classes. See more analyses in Sections~\ref{sec:misalignment} and~\ref{sec:more_analyses}.

    \item We propose a novel model, MARS, to tackle these challenges. Specifically, we design Boundary-Preserving Alignment to mitigate boundary erosion by exploiting the visual structure of zero-shot CLIP, and develop Margin-aware Self-Refinement to adaptively balance teacher guidance and pseudo-label supervision, further improving discriminability for tail and confusable classes.

    \item Comprehensive experiments on nine benchmarks with different imbalance ratios show that MARS achieves state-of-the-art performance with an average improvement of \textbf{4.71} percentage points.
\end{itemize}

\section{Related Work}
\label{sec:related}

\subsection{Prompt Learning in VLMs}

Vision-language models (VLMs)~\citep{radford2021clip, jia2021align, li2022blip, xiao2024florence2} have achieved remarkable zero-shot generalization by aligning visual and textual modalities. Prompt learning is a parameter-efficient paradigm for adapting these models to downstream tasks. Specifically, CoOp~\citep{zhou2022coop, zhou2022cocoop} pioneers this direction by optimizing context vectors within the text encoder. VPT~\citep{jia2022vpt} extends this concept to the visual modality by injecting learnable prompts into the image encoder. To facilitate cross-modal interaction, MaPLe~\citep{khattak2023maple} introduces a multi-modal prompting approach to ensure deep alignment between branches. Alternatively, some works focus on refining the encoder's output features without fine-tuning the backbone. For example, CLIP-Adapter~\citep{gao2024adapter} employs lightweight residual modules to transform visual and textual features.
\textbf{Despite their success, these methods heavily rely on labeled data, which severely constrains their applicability in unsupervised scenarios.}

\subsection{Learning from Unlabeled Data for VLMs}

To mitigate the reliance on labeled data, various strategies leverage the zero-shot capabilities of VLMs to generate pseudo-labels for self-training. For instance, UPL~\citep{huang2022upt} and GRIP~\citep{menghini2023grip} focus on improving pseudo-label quality via confidence-based filtering or iterative top-$k$ selection. Similarly, CPL~\citep{zhang2024cpl} assigns each sample a set of candidate pseudo-labels at each iteration, expecting the true label to be captured within the set. Beyond traditional pseudo-labeling, recent studies explore novel perspectives for mining supervisory signals from unlabeled data. UEO~\citep{liang2024ueo} employs marginal entropy as a unified metric to regularize the decision boundary, minimizing uncertainty for confident predictions while maximizing it for ambiguous samples. TMP~\citep{li2025tmp} reformulates the adaptation process as a binary verification task instead of the traditional classification task. CAP~\citep{wang2025cap} identifies inherent biases and concept confusion in CLIP zero-shot predictions, rectifying these issues by learning a projection layer that encourages balanced predictions across classes. microCLIP~\citep{silva2026microclip} combines coarse-to-fine token fusion and saliency-aware pooling for unsupervised CLIP adaptation.
\textbf{However, their reliance on a uniform prior over unlabeled data limits their applicability to long-tailed scenarios.}

\section{Methodology}

\subsection{Problem Formulation and Challenges}
\label{sec:3.1}

In unsupervised long-tailed adaptation (ULTA), we consider an unlabeled dataset $\mathcal{D}_u = \{x_i\}_{i=1}^N$ of size $N$ belonging to $C$ classes. While one can theoretically define $N_c$ as the number of samples for class $c$ and $\gamma = \max_c N_c / \min_c N_c$ as the imbalance ratio, neither is accessible in practice. The core challenge is that existing methods implicitly assume a uniform prior when generating pseudo-labels, which inevitably leads to severe boundary erosion and prediction bias. Our goal is to achieve robust performance across different imbalance ratios.

\subsection{Analysis of Misalignment}
\label{sec:misalignment}

Existing unsupervised methods typically rely on a uniform prior to generate pseudo-labels. In long-tailed scenarios, this assumption forces the model to align visual features with incorrect targets, causing a severe misalignment between the learned features and the true semantic information. We analyze the consequences of this misalignment from the perspective of gradient dynamics.

Consider a head-class sample $x$ with normalized visual feature $\mathbf{v} \in \mathbb{R}^d$ and ground-truth label $c_h$, which is incorrectly assigned a tail-class pseudo-label $\hat{y}=c_t$.
In the standard self-training framework, this incorrect pseudo-label is treated as a hard target, forcing the visual feature to align with the textual embedding $\mathbf{t}_{c_t}$.
Formally, we minimize the cross-entropy loss $\mathcal{L}_{ce} = -\log p(c_t|\mathbf{v})$, where the probability $p$ is computed via cosine similarity with textual embeddings $\{\mathbf{t}_c\}_{c=1}^C$ scaled by a temperature parameter $\tau$. The gradient of this loss with respect to the visual feature $\mathbf{v}$ is derived as:
\begin{equation}
    \nabla_{\mathbf{v}} \mathcal{L}_{ce} = \frac{1}{\tau} \left( \sum_{j=1}^C p(j|\mathbf{v}) \cdot \mathbf{t}_j - \mathbf{t}_{c_t} \right).
\end{equation}
Before fitting this incorrect pseudo-label, the model still assigns high confidence to the correct head class $c_h$, so the probability mass is concentrated on $c_h$.
Consequently, the gradient approximates:
\begin{equation}
    \nabla_{\mathbf{v}} \mathcal{L}_{ce} \approx \frac{1}{\tau} (\mathbf{t}_{c_h} - \mathbf{t}_{c_t}).
    \label{eq:grad_approx}
\end{equation}

\begin{figure}[t!]
    \centering

    \begin{minipage}{1\linewidth}
        \centering

        \includegraphics[width=0.25\linewidth]{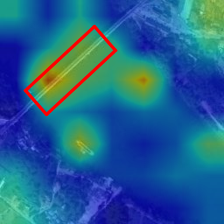}
        \hfill
        \includegraphics[width=0.25\linewidth]{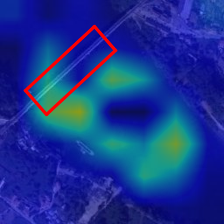}
        \hfill
        \includegraphics[width=0.25\linewidth]{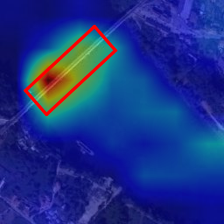}

        \vspace{0cm}

        \parbox[t]{0.25\linewidth}{
            \phantomsubcaption\label{fig:3_a}
            \centering \small (a) Zero-shot CLIP
        }
        \hfill
        \parbox[t]{0.25\linewidth}{
            \phantomsubcaption\label{fig:3_b}
            \centering \small (b) Baseline
        }
        \hfill
        \parbox[t]{0.25\linewidth}{
            \phantomsubcaption\label{fig:3_c}
            \centering \small (c) Ours
        }

    \end{minipage}

    \vspace{-0.15cm}

    \caption{
    Visualization of attention maps for misassigned samples on RESISC45.
    We select head-class samples where the baseline method incorrectly assigns tail-class pseudo-labels.
    \textbf{(a)} The model mainly focuses on the foreground object.
    \textbf{(b)} The model is distracted by the irrelevant background.
    \textbf{(c)} The model correctly captures the foreground object while reducing background attention.
    }
    \label{fig:3}

    \vspace{-0.2cm}
\end{figure}

The feature vector $\mathbf{v}$ is updated by moving against the gradient direction (i.e., $\mathbf{v} \leftarrow \mathbf{v} - \eta \nabla_{\mathbf{v}} \mathcal{L}_{ce}$). Consequently, the resulting feature shift satisfies $\Delta \mathbf{v} \propto \mathbf{t}_{c_t} - \mathbf{t}_{c_h}$. This derivation illustrates that the gradient drives the visual feature $\mathbf{v}$ to align with the distant tail embedding $\mathbf{t}_{c_t}$, deviating from its true semantic embedding $\mathbf{t}_{c_h}$ and thereby eroding the decision boundaries of head classes. Meanwhile, to minimize the loss, the model is forced to exploit irrelevant background noise. As visualized in Fig.~\ref{fig:3_b}, for these mislabeled samples, existing methods (e.g., CPL~\citealp{zhang2024cpl}) cause the model to ignore the foreground object and focus on the background, thereby degrading feature discriminability.

\subsection{Boundary-Preserving Alignment}
\label{sec:alignment}

To mitigate this misalignment, we prevent erroneous probability changes from becoming training targets. In self-training, the targets are generated from the model's own predictions, so an erroneous prediction is fitted again in the next round. We therefore take zero-shot CLIP, which remains fixed during adaptation, as a visual reference, and correct the targets according to whether each probability increase is supported by visual evidence.

Concretely, let $x_i^w$ denote the weak view of $x_i$. We extract the unit-normalized feature $\mathbf{u}_i$ and the predicted distribution $a_i$ of zero-shot CLIP on $x_i^w$, and denote the zero-shot class as $c_i=\arg\max_c a_{ic}$. For each class $c$, we average the features of the samples with $c_i=c$ and $\ell_2$-normalize the result to obtain the class-level visual anchor $\boldsymbol{\mu}_c$, and measure the visual support of sample $i$ for class $c$ by $s_{ic}=\mathbf{u}_i^\top\boldsymbol{\mu}_c$. These anchors characterize the class-level visual structure of zero-shot CLIP. Unlike the textual embeddings, which always favor the zero-shot class, they compare each sample with the visual centroids of the zero-shot groups and thus provide complementary evidence.

At the start of each alignment epoch, let $p_i$ denote the predicted distribution of the current model on $x_i^w$ and $b_i=\arg\max_c p_{ic}$. For $j\ne c_i$, the probability increase induced by adaptation is $g_{ij}=\max(p_{ij}-a_{ij},0)$. We regard this increase as visually unsupported, i.e., $h_{ij}=1$, if $s_{ic_i}>s_{ij}$, and set $h_{ij}=0$ otherwise. If $b_i\ne c_i$ and $s_{ib_i}>s_{ic_i}$, the adaptation has likely corrected a zero-shot error, and we set $h_{ij}=0$ for all $j$.

Since the verdict of a single sample is noisy and pseudo-label errors tend to recur between the same class pairs (Fig.~\ref{fig:2_b}), we aggregate the evidence over the unlabeled data:
\begin{equation}
    \rho_{cj}=
    \frac{\sum_{i:c_i=c} g_{ij}h_{ij}}
         {\sum_{i:c_i=c} g_{ij}},\qquad c\ne j.
    \label{eq:bpa_direction}
\end{equation}
The corrected target then returns the unsupported part of each increase to the zero-shot class:
\begin{equation}
    q_i=p_i+\sum_{j\ne c_i}\delta_{ij}(\mathbf e_{c_i}-\mathbf e_j),
    \qquad
    \delta_{ij}=\rho_{c_i j}\,h_{ij}\,g_{ij},
    \label{eq:bpa_target}
\end{equation}
where $\mathbf e_c$ is the one-hot vector for class $c$. Since $0\le\delta_{ij}\le g_{ij}\le p_{ij}$, $q_i$ remains a valid distribution. We fix $q_i$ within each epoch and use it to supervise the strong view $x_i^s$ generated by RandAugment~\citep{cubuk2020randaugment}:
\begin{equation}
    \mathcal L_{\mathrm{BPA}}=
    -\frac1B\sum_{i\in\mathcal B}\sum_{c=1}^{C}
    q_{ic}\log p(c\mid x_i^s),
    \label{eq:bpa_loss}
\end{equation}
where $p(c\mid x_i^s)$ is the predicted probability of class $c$ on $x_i^s$ and $\mathcal B$ is a minibatch of size $B$.

Crucially, let $\ell(q_i)$ denote the cross-entropy between target $q_i$ and the softmax of the strong-view logits $z_i$. Compared with the uncorrected target $p_i$, the corrected target changes the gradient by
\begin{equation}
    \nabla_{z_i}\ell(q_i)-\nabla_{z_i}\ell(p_i)
    =-\sum_{j\ne c_i}\delta_{ij}
       (\mathbf e_{c_i}-\mathbf e_j).
    \label{eq:bpa_gradient}
\end{equation}
Each unsupported increase thus induces a push along $\mathbf e_{c_i}-\mathbf e_j$ on the logits. Since $z_{ic}=\mathbf v^\top\mathbf t_c/\tau$ for the visual feature $\mathbf v$ of $x_i^s$, this push moves $\mathbf v$ toward $\mathbf t_{c_i}$ and away from $\mathbf t_j$, counteracting the misaligned gradient in Eq.~\eqref{eq:grad_approx}, while supported increases, including those that correct zero-shot errors, are preserved. As shown in Fig.~\ref{fig:3_c}, BPA restores the attention to the foreground object, and Fig.~\ref{fig:4}\subref{fig:4_a} shows that it keeps head-class accuracy stable as the imbalance ratio increases.

\subsection{Margin-Aware Self-Refinement}
\label{sec:refinement}

Although boundary-preserving alignment establishes a foundation in the joint feature space, refining the decision boundaries for the tail and confusable classes remains a critical challenge in unsupervised long-tailed scenarios. To address this, instead of directly fitting noisy pseudo-labels that can destroy the boundary, we propose a margin-aware self-refinement strategy that improves the feature space by decoupling well-learned boundary preservation from targeted feature enhancement.

We employ the aligned model from Section~\ref{sec:alignment} as a frozen teacher and introduce a learnable adapter as the student. The refinement objective is formulated as follows:
\begin{equation}
    \mathcal{L}_{\text{refine}} = \alpha \cdot \mathcal{L}_{\text{KL}}(p^T, p^S) + (1 - \alpha) \cdot \mathcal{L}_{\text{CE}}(\hat{y}, p^S),
\end{equation}
where $p^T$ and $p^S$ denote the predictions of the frozen teacher and the student, respectively. In this formulation, the Kullback-Leibler (KL) divergence loss $\mathcal{L}_{\text{KL}}$ serves as a safety anchor to preserve the well-learned concepts from the frozen teacher, while the standard cross-entropy loss $\mathcal{L}_{\text{CE}}$ leverages pseudo-label supervision with $\hat y=\arg\max_c p_c^T$ to acquire new knowledge. The gating factor $\alpha\in[0,1]$ plays a critical role in balancing knowledge preservation against the learning of new features.

\begin{figure*}[t]
    \centering
    \includegraphics[width=0.98\textwidth]{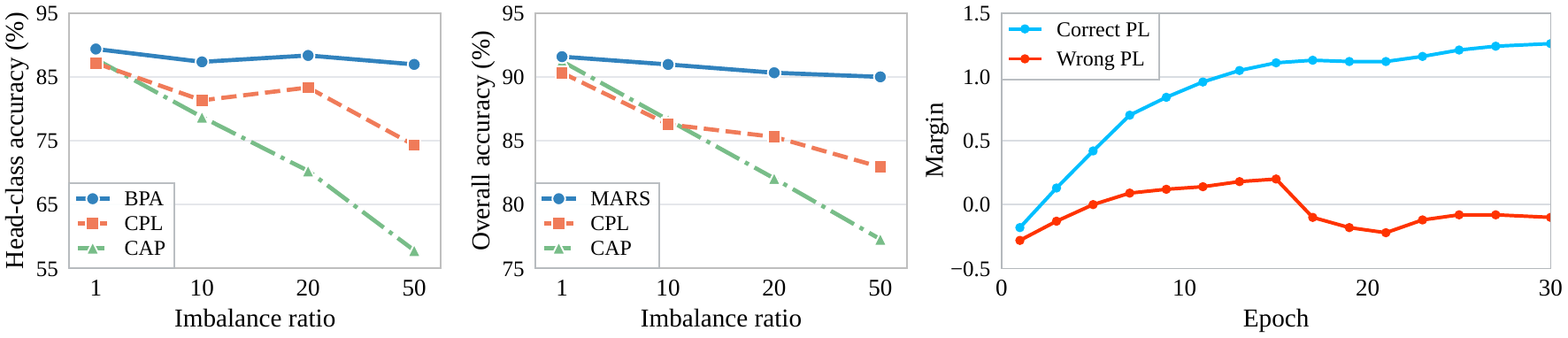}
    \vspace{-0.08cm}
    \renewcommand{\thesubfigure}{\alph{subfigure}}
    \captionsetup[subfigure]{labelfont={normalfont},textfont={normalfont},skip=0pt}
    \parbox[t]{0.31\textwidth}{%
        \phantomsubcaption\label{fig:4_a}%
        \vspace{-0.2cm}
        \centering \small (a)%
    }
    \hfill
    \parbox[t]{0.18\textwidth}{%
        \phantomsubcaption\label{fig:4_b}%
        \vspace{-0.2cm}
        \centering \small (b)%
    }
    \hfill
    \parbox[t]{0.38\textwidth}{%
        \phantomsubcaption\label{fig:4_c}%
        \vspace{-0.2cm}
        \centering \small (c)%
    }

    \vspace{-0.1cm}

    \caption{\textbf{(a)} BPA preserves head-class accuracy on OxfordPets as the imbalance ratio increases. \textbf{(b)} The complete MARS pipeline maintains stable overall accuracy on OxfordPets across imbalance ratios. \textbf{(c)} Evolution of classification margins during training. Samples with correct pseudo-labels (Correct PL) consistently exhibit increasing margins, whereas mislabeled samples (Wrong PL) fluctuate around zero or remain negative.}
    \label{fig:4}
    \vspace{-0.15cm}
\end{figure*}

To adaptively modulate $\alpha$, we introduce a class-level prior. Specifically, we define the gating factor as $\alpha = (r_{\hat{y}})^\beta$, where $\beta$ is a hyperparameter.
We fix $\beta=1$ across all datasets, and the sensitivity analysis in Appendix~\ref{app:sensitivity_hyperparameters} further shows that our method is robust to this hyperparameter.
Here, $r_{\hat{y}} = \frac{\hat{n}_{\hat{y}}}{\max_{c}(\hat{n}_c)}$ represents the normalized frequency, where $\hat{n}_{\hat{y}}$ denotes the estimated sample count for class $\hat{y}$ accumulated from the teacher's soft predictions. High-frequency predictions primarily correspond to head classes, which require a higher $\alpha$ to maintain the model's well-established boundary. In contrast, low-frequency predictions correspond to tail and confusable classes, which necessitate a lower $\alpha$ to facilitate the acquisition of new knowledge, thereby enhancing feature discriminability.

However, relying solely on class-level priors is insufficient, as it overlooks the quality of individual pseudo-labels within the same class. As illustrated in Fig.~\ref{fig:4}\subref{fig:4_c}, our empirical observations reveal that correctly pseudo-labeled samples mostly exhibit increasing margins during training, whereas incorrectly pseudo-labeled samples maintain negative margins. Motivated by this, we propose Cumulative Margin Consistency (CMC) to quantify sample quality. Specifically, given the student's logits $z^{(t)}$ at epoch $t$, the margin is defined as:
\begin{equation}
    M^{(t)}(x, \hat{y}) = z_{\hat{y}}^{(t)}(x) - \max_{j \neq \hat{y}} z_{j}^{(t)}(x).
\end{equation}
We calculate the exponential moving average of this margin with momentum $\lambda$ as $\text{CMC}(x, \hat{y})$ and refine $\alpha$ into a sample-specific factor $\hat{\alpha}$ to mitigate the impact of noisy pseudo-labels:
\begin{equation}
    \hat{\alpha} = \frac{1}{2}\left(
    r_{\hat{y}}^{\beta} + (1 - \mathcal{N}(\text{CMC}(x, \hat{y})))
    \right),
    \label{eq:alpha}
\end{equation}
where $\mathcal{N}(\cdot)$ is the min-max normalization function. The final refinement objective is formulated as:
\begin{equation}
    \mathcal{L}_{\text{refine}} = \hat{\alpha} \cdot \mathcal{L}_{\text{KL}}(p^T, p^S) + (1 - \hat{\alpha}) \cdot \mathcal{L}_{\text{CE}}(\hat{y}, p^S).
    \label{eq:refine_loss}
\end{equation}

The resulting stability across imbalance ratios is reflected in Fig.~\ref{fig:4}\subref{fig:4_b}: the complete MARS pipeline retains its overall accuracy as the long-tailed distribution becomes more severe.

\subsection{Model Training and Prediction}
\label{sec:training_prediction}

Our model consists of Boundary-Preserving Alignment (BPA) and Margin-aware Self-Refinement (MSR).
The training process first performs boundary-preserving alignment with corrected weak-view targets, and then conducts margin-aware self-refinement with the frozen teacher to selectively refine the remaining hard classes and samples.
The detailed training procedure is provided in Appendix~\ref{app:training_process}.
Notably, both the teacher and student adapters are appended to the textual encoder. The pre-trained image and text encoders are kept frozen, and only the prompts and adapters are learned.

Upon completion of training, the teacher adapter is discarded.
For an unseen test sample $x^*$, the predicted label $y^*$ is obtained by computing the cosine similarity between the visual feature $\mathbf{v}^*$ extracted by the refined encoder and the pre-computed textual embeddings $\{\mathbf{t}_c\}_{c=1}^C$:
\begin{equation}
    y^* = \operatorname*{arg\,max}_{c \in \{1, \dots, C\}} \left( \frac{\mathbf{v}^* \cdot \mathbf{t}_c}{\tau} \right).
\end{equation}

\FloatBarrier

\begin{table}[!t]
\captionsetup{skip=0pt}
\caption{Comparison of the test accuracy (\%) across three seeds on nine benchmark datasets with different imbalance ratios (10, 20, 50). The results are reported as $\text{mean}_{\text{std}}$ over three random seeds, with the best results presented in bold. Zero-shot CLIP is abbreviated as ZS CLIP.}
\label{tab:main_results}
\vspace{0.15cm}
\centering
\resizebox{\linewidth}{!}{
\setlength{\tabcolsep}{4pt}
\begin{tabular}{l|ccccccccc|c}
\toprule

\multicolumn{11}{c}{\textbf{Imbalance Ratio = 10}} \\
\midrule
Method & CUB. & Res. & FA. & Cal. & ES. & Flw. & Food. & Pets. & Cars. & Avg. \\
\midrule
ZS CLIP & \res{51.82}{0.00} & \res{54.48}{0.00} & \res{17.58}{0.00} & \res{90.30}{0.00} & \res{32.88}{0.00} & \res{63.67}{0.00} & \res{78.80}{0.00} & \res{84.36}{0.00} & \res{58.16}{0.00} & \res{59.12}{0.00} \\
\midrule
 FPL (NeurIPS'23) & \res{51.59}{0.23} & \res{62.58}{0.86} & \res{19.05}{0.37} & \res{90.67}{0.39} & \res{55.49}{1.35} & \res{61.30}{1.72} & \res{78.10}{0.15} & \res{86.44}{0.33} & \res{58.10}{0.28} & \res{62.59}{0.63} \\
 GRIP (NeurIPS'23) & \res{48.01}{0.68} & \res{61.95}{0.99} & \res{18.25}{0.24} & \res{88.98}{0.62} & \res{58.44}{5.52} & \res{64.00}{1.55} & \res{73.06}{0.49} & \res{84.15}{0.29} & \res{50.93}{0.36} & \res{60.86}{1.19} \\
CPL (ICML'24) & \res{49.27}{0.69} & \res{61.42}{1.22} & \res{17.58}{0.32} & \res{82.77}{0.37} & \res{63.49}{6.24} & \res{61.91}{1.04} & \res{72.11}{0.23} & \res{86.28}{0.30} & \res{49.31}{0.59} & \res{60.46}{1.22} \\
UEO (ICML'24) & \res{53.56}{0.11} & \res{60.07}{0.01} & \res{18.73}{0.15} & \res{92.74}{0.08} & \res{43.68}{0.14} & \res{65.71}{0.07} & \res{80.67}{0.04} & \res{87.94}{0.09} & \res{58.03}{0.18} & \res{62.35}{0.09} \\
TMP (ICML'25) & \res{18.49}{0.18} & \res{39.78}{0.26} & \res{10.97}{0.12} & \res{82.76}{0.26} & \res{16.80}{0.30} & \res{13.16}{0.29} & \res{59.64}{0.65} & \res{33.96}{3.59} & \res{19.46}{0.10} & \res{32.78}{0.64} \\
CAP (ICML'25) & \res{51.75}{0.55} & \res{68.83}{2.66} & \res{16.02}{2.07} & \res{89.72}{3.51} & \res{66.37}{2.67} & \res{60.88}{1.73} & \res{80.68}{0.37} & \res{86.64}{1.41} & \res{58.26}{0.43} & \res{64.35}{1.71} \\
microCLIP (ACL'26) & \res{49.88}{0.59} & \res{67.98}{0.87} & \res{17.28}{0.87} & \res{93.01}{0.12} & \res{51.76}{1.29} & \res{67.96}{1.34} & \res{79.94}{0.15} & \res{88.77}{0.94} & \res{58.94}{0.34} & \res{63.95}{0.24} \\
\midrule
\rowcolor[HTML]{E8E5E5} \algo~(Ours) & \res{\textbf{54.84}}{\textbf{0.35}} & \res{\textbf{68.85}}{\textbf{0.41}} & \res{\textbf{20.24}}{\textbf{0.27}} & \res{\textbf{94.06}}{\textbf{0.51}} & \res{\textbf{74.37}}{\textbf{0.29}} & \res{\textbf{68.95}}{\textbf{1.00}} & \res{\textbf{81.52}}{\textbf{0.17}} & \res{\textbf{90.97}}{\textbf{0.42}} & \res{\textbf{62.38}}{\textbf{0.52}} & \res{\textbf{68.47}}{\textbf{0.20}} \\

\midrule
\multicolumn{11}{c}{\textbf{Imbalance Ratio = 20}} \\
\midrule
Method & CUB. & Res. & FA. & Cal. & ES. & Flw. & Food. & Pets. & Cars. & Avg. \\
\midrule
ZS CLIP & \res{51.82}{0.00} & \res{54.48}{0.00} & \res{17.58}{0.00} & \res{90.30}{0.00} & \res{32.88}{0.00} & \res{63.67}{0.00} & \res{78.80}{0.00} & \res{84.36}{0.00} & \res{58.16}{0.00} & \res{59.12}{0.00} \\
\midrule
FPL (NeurIPS'23) & \res{51.04}{0.26} & \res{60.03}{0.75} & \res{18.46}{0.21} & \res{90.30}{0.09} & \res{51.89}{1.32} & \res{60.87}{1.52} & \res{77.48}{0.15} & \res{86.77}{0.33} & \res{57.51}{0.41} & \res{61.60}{0.56} \\
GRIP (NeurIPS'23) & \res{47.63}{0.75} & \res{59.27}{1.74} & \res{17.20}{0.20} & \res{87.75}{0.18} & \res{60.01}{2.57} & \res{62.26}{0.91} & \res{70.86}{0.27} & \res{81.85}{0.67} & \res{49.65}{0.93} & \res{59.61}{0.91} \\
CPL (ICML'24) & \res{48.46}{0.39} & \res{58.45}{0.32} & \res{16.81}{0.23} & \res{79.99}{3.02} & \res{64.35}{0.34} & \res{60.63}{0.94} & \res{70.14}{0.30} & \res{85.30}{0.31} & \res{47.83}{0.61} & \res{59.11}{0.72} \\
UEO (ICML'24) & \res{53.20}{0.09} & \res{58.69}{0.11} & \res{17.28}{1.01} & \res{92.75}{0.06} & \res{43.15}{0.03} & \res{66.13}{0.04} & \res{80.48}{0.09} & \res{87.61}{0.10} & \res{57.82}{0.12} & \res{61.90}{0.18} \\
TMP (ICML'25) & \res{15.61}{0.52} & \res{32.71}{0.73} & \res{8.10}{0.05} & \res{69.82}{0.67} & \res{12.27}{1.69} & \res{9.51}{0.62} & \res{50.46}{0.45} & \res{22.03}{1.43} & \res{15.11}{0.14} & \res{26.18}{0.70} \\
 CAP (ICML'25) & \res{49.89}{0.57} & \res{59.94}{3.42} & \res{13.47}{1.82} & \res{88.15}{4.36} & \res{62.07}{0.75} & \res{59.96}{0.62} & \res{78.88}{1.49} & \res{82.05}{3.28} & \res{56.08}{0.49} & \res{61.17}{1.87} \\
microCLIP (ACL'26) & \res{49.11}{0.38} & \res{66.47}{0.58} & \res{16.70}{0.23} & \res{92.89}{0.20} & \res{52.65}{4.81} & \res{66.53}{1.55} & \res{79.46}{0.19} & \res{87.98}{0.75} & \res{56.89}{0.78} & \res{63.19}{0.54} \\
\midrule
\rowcolor[HTML]{E8E5E5} \algo~(Ours) & \res{\textbf{53.80}}{\textbf{0.82}} & \res{\textbf{68.42}}{\textbf{0.58}} & \res{\textbf{20.45}}{\textbf{0.84}} & \res{\textbf{94.04}}{\textbf{0.15}} & \res{\textbf{72.17}}{\textbf{3.73}} & \res{\textbf{67.30}}{\textbf{0.86}} & \res{\textbf{81.52}}{\textbf{0.26}} & \res{\textbf{90.32}}{\textbf{0.26}} & \res{\textbf{61.18}}{\textbf{0.86}} & \res{\textbf{67.69}}{\textbf{0.52}} \\

\midrule
\multicolumn{11}{c}{\textbf{Imbalance Ratio = 50}} \\
\midrule
Method & CUB. & Res. & FA. & Cal. & ES. & Flw. & Food. & Pets. & Cars. & Avg. \\
\midrule
ZS CLIP & \res{51.82}{0.00} & \res{54.48}{0.00} & \res{17.58}{0.00} & \res{90.30}{0.00} & \res{32.88}{0.00} & \res{63.67}{0.00} & \res{78.80}{0.00} & \res{84.36}{0.00} & \res{58.16}{0.00} & \res{59.12}{0.00} \\
\midrule
FPL (NeurIPS'23) & \res{50.60}{0.32} & \res{58.07}{0.51} & \res{17.95}{0.66} & \res{89.67}{0.75} & \res{49.53}{2.32} & \res{60.09}{2.26} & \res{76.78}{0.35} & \res{86.41}{0.54} & \res{56.76}{0.29} & \res{60.65}{0.89} \\
GRIP (NeurIPS'23) & \res{47.57}{0.28} & \res{56.48}{0.98} & \res{16.47}{0.24} & \res{86.52}{0.47} & \res{53.84}{3.69} & \res{60.99}{1.23} & \res{68.99}{0.73} & \res{81.14}{0.80} & \res{49.26}{0.84} & \res{57.92}{1.03} \\
CPL (ICML'24) & \res{47.88}{0.43} & \res{56.64}{0.28} & \res{16.43}{0.23} & \res{77.73}{2.75} & \res{59.76}{6.11} & \res{60.58}{0.93} & \res{68.24}{0.19} & \res{82.94}{0.10} & \res{46.71}{0.80} & \res{57.43}{1.31} \\
UEO (ICML'24) & \res{52.46}{0.06} & \res{56.96}{0.17} & \res{17.49}{0.52} & \res{92.77}{0.06} & \res{42.37}{0.05} & \res{66.01}{0.07} & \res{80.24}{0.05} & \res{86.99}{0.03} & \res{57.23}{0.17} & \res{61.39}{0.13} \\
TMP (ICML'25) & \res{12.86}{0.05} & \res{26.21}{0.12} & \res{5.53}{0.37} & \res{60.45}{0.74} & \res{11.33}{1.64} & \res{8.21}{0.25} & \res{38.63}{0.52} & \res{14.31}{1.35} & \res{12.49}{0.29} & \res{21.11}{0.59} \\
CAP (ICML'25) & \res{49.11}{0.27} & \res{57.29}{1.31} & \res{12.77}{1.70} & \res{86.66}{5.99} & \res{52.88}{1.60} & \res{58.43}{0.68} & \res{77.89}{3.03} & \res{77.29}{5.61} & \res{53.10}{0.30} & \res{58.38}{2.28} \\
microCLIP (ACL'26) & \res{48.11}{0.60} & \res{63.73}{0.79} & \res{15.36}{0.42} & \res{92.25}{0.51} & \res{48.16}{3.79} & \res{64.88}{0.68} & \res{79.22}{0.29} & \res{87.93}{0.43} & \res{55.47}{0.83} & \res{61.68}{0.53} \\
\midrule
\rowcolor[HTML]{E8E5E5} \algo~(Ours) & \res{\textbf{54.29}}{\textbf{0.68}} & \res{\textbf{68.16}}{\textbf{0.47}} & \res{\textbf{20.05}}{\textbf{0.26}} & \res{\textbf{93.18}}{\textbf{0.08}} & \res{\textbf{69.22}}{\textbf{5.60}} & \res{\textbf{68.13}}{\textbf{1.00}} & \res{\textbf{81.24}}{\textbf{0.47}} & \res{\textbf{90.00}}{\textbf{0.79}} & \res{\textbf{60.55}}{\textbf{0.46}} & \res{\textbf{67.20}}{\textbf{0.90}} \\
\bottomrule
\end{tabular}
}
\vspace{-0.2cm}
\end{table}

\section{Experiments}
\label{sec:experiments}

\subsection{Experimental Setting}
\label{sec:experimental_setting}

\textbf{Datasets.}
We conduct extensive experiments on nine classification datasets from diverse domains, including CUB (CUB.; ~\citealp{wah2011cub}), RESISC45 (Res.; ~\citealp{cheng2017res}), FGVCAircraft (FA.; ~\citealp{maji2013fa}), Caltech101 (Cal.; ~\citealp{fei2004cal}), EuroSAT (ES.; ~\citealp{2019es}), Flowers102 (Flw.; ~\citealp{2008flw}), Food101 (Food.; ~\citealp{bossard2014food}), OxfordPets (Pets.; ~\citealp{2012pets}), and StanfordCars (Cars.; ~\citealp{fei2013cars}).
For the long-tailed settings, the training data is generated by down-sampling the original datasets to follow an exponential decay of different ratios.
Specifically, the maximum number of samples per class is set to 100 (or the original class size), while the minimum is constrained to 2.
Following the definition in Section~\ref{sec:3.1}, we set the imbalance ratio $\gamma \in \{10, 20, 50\}$ to simulate different long-tailed scenarios.
For the fine-grained analysis, classes are divided into Head, Medium, and Tail groups using a 15/30/55 split by training-set frequency.
The details of dataset statistics can be found in Appendix~\ref{app:datasets_details}.

\textbf{Baselines.}
We compare our method with seven existing methods, namely,
Few Pseudo-Labels (FPL; ~\citealp{menghini2023grip}), Grow and Refine Iteratively Pseudo-labels (GRIP; ~\citealp{menghini2023grip}), Candidate Pseudo-Labeling (CPL; ~\citealp{zhang2024cpl}), Universal Entropy Optimization (UEO; ~\citealp{liang2024ueo}), True-false labels via Multimodal Prompt retrieving (TMP; ~\citealp{li2025tmp}), Concept-Adaptive Pseudo-labeling (CAP; ~\citealp{wang2025cap}), and Coarse-Fine Token Fusion (microCLIP; ~\citealp{silva2026microclip}).
For fair comparison, we standardize the visual backbone to OpenAI CLIP ViT-B/32 and use the same training dataset and evaluation metrics.
All other settings follow the official recommendations.

\textbf{Implementation details.}
Following the standard protocol, we set the multi-modal prompts as MaPLe~\citep{khattak2023maple}, with a depth of 9 and a context length of 2, initialized with the template ``a photo of a {}''.
The weak view follows the standard CLIP preprocessing, while the strong view is generated by RandAugment~\citep{cubuk2020randaugment}.
We fix $\beta=1$ and the momentum coefficient ($\lambda=0.95$) across all datasets.
For a comprehensive evaluation of different VLMs, we also report results using OpenCLIP ViT-B/32 (LAION-2B)~\citep{cherti2023openclip} and SigLIP ViT-B/16~\citep{zhai2023siglip}.
Additional reproduction details can be found in Appendix~\ref{app:implementation_details}.

\subsection{Main Results}

\textbf{Our method achieves SOTA classification accuracy.}
Table~\ref{tab:main_results} demonstrates that our method outperforms both zero-shot CLIP and existing unsupervised methods by a substantial margin across nine benchmark datasets under varying imbalance ratios (i.e., 10, 20, 50).
Specifically, the performance of our proposed method exceeds the previous state-of-the-art method by an average of \textbf{4.12}, \textbf{4.50}, and \textbf{5.52} percentage points (pp) across three seeds under imbalance ratios of 10, 20, and 50, respectively.
In the most challenging scenario with an imbalance ratio of 50, our method demonstrates remarkable robustness. While CAP suffers a 5.97~pp drop as the ratio increases from 10 to 50, our method exhibits high stability with a minimal decline of only 1.27~pp.
Notably, at an imbalance ratio of 50, four of the seven baselines even fall below zero-shot CLIP on average, whereas MARS still improves over it by \textbf{8.08}~pp.
Moreover, under an imbalance ratio of 50, our method improves RESISC45 accuracy over the best existing method by \textbf{4.43}~pp.
Furthermore, pairwise t-tests at a 0.05 significance level using three corresponding seed results show that our method achieves significant improvements in \textbf{84.7\%} of cases (160 out of 189), further confirming its effectiveness in ULTA. Detailed results can be found in Appendix~\ref{app:statistical_significance}.

\setlength{\columnsep}{0.25cm}

\begin{wraptable}[11]{r}{0.45\linewidth}
\centering
\vspace{-0.46cm}

\captionsetup{skip=2pt}
\caption{Comparison of test accuracy (\%) under the balanced setting.}
\label{tab:balanced}

\vspace{0.1cm}

\resizebox{\linewidth}{!}{
    \setlength{\tabcolsep}{7pt}
    \begin{tabular}{l|ccc|c}
    \toprule
    Method & Cal. & Flw. & Pets. & Avg. \\
    \midrule
    ZS CLIP  & 90.30 & 63.67 & 84.36 & 79.44 \\
    \midrule
    FPL  & 92.08 & 67.47 & 87.41 & 82.32 \\
    CPL  & 92.37 & 68.20 & 90.32 & 83.63 \\
    UEO  & 92.98 & 66.82 & 89.13 & 82.98 \\
    CAP  & 94.36 & 72.35 & 91.22 & 85.98 \\
    microCLIP & 94.12 & 72.11 & 89.92 & 85.38 \\
    \midrule
    \rowcolor[HTML]{E8E5E5} \algo
    & \textbf{94.89} & \textbf{72.35} & \textbf{91.58} & \textbf{86.27} \\
    \bottomrule
    \end{tabular}
}

\vspace{-0.35cm}
\end{wraptable}

\textbf{Our method maintains performance stability and robustness under a uniform distribution.}
Using the same comparison protocol as Section~\ref{sec:experimental_setting}, we further evaluate our method on three datasets (Caltech101, Flowers102, and OxfordPets) in a uniform setting.
As reported in Table~\ref{tab:balanced}, MARS outperforms existing unsupervised methods.
Specifically, MARS achieves the highest average accuracy of 86.27\%, surpassing the leading baseline by 0.30~pp.
This validates that our method enhances feature discriminability without compromising robustness, ensuring reliable performance across unknown distributions.

\setlength{\columnsep}{0.25cm}
\begin{wraptable}{r}{0.45\linewidth}
\centering
\vspace{-0.45cm}

\captionsetup{skip=2pt}
\caption{Test accuracy (\%) using OpenCLIP with an imbalance ratio of 20. Zero-shot OpenCLIP is abbreviated as ZS OCLIP.}
\label{tab:foundation_model}

\vspace{0cm}

\resizebox{\linewidth}{!}{
    \setlength{\tabcolsep}{6pt}
    \begin{tabular}{l|cccc|c}
    \toprule
    Method & Res. & Flw. & ES. & Pets. & Avg. \\
    \midrule
    ZS OCLIP & 63.90 & 71.43 & 51.74 & 90.62 & 69.42 \\
    \midrule
    FPL & 60.59 & 66.04 & 50.32 & 87.54 & 66.12 \\
    GRIP & 63.25 & 62.08 & 61.24 & 80.84 & 66.85 \\
    CPL & 62.68 & 63.41 & 64.44 & 84.60 & 68.78 \\
    UEO & 59.57 & 70.35 & 52.06 & 88.88 & 67.72 \\
    TMP & 41.07 & 13.60 & 10.00 & 47.75 & 28.11 \\
    CAP & 63.25 & 61.68 & 63.30 & 80.54 & 67.19 \\
    microCLIP & 66.55 & 66.16 & 58.56 & 89.21 & 70.12 \\
    \midrule
    \rowcolor[HTML]{E8E5E5} \algo & \textbf{73.12} & \textbf{74.22} & \textbf{80.88} & \textbf{92.40} & \textbf{80.16} \\
    \bottomrule
    \end{tabular}
}

\vspace{-0.3cm}
\end{wraptable}
\textbf{Our method is effective across different VLMs.}
To verify the universality of our method, we extend our evaluation to different vision-language models.
As illustrated in Table~\ref{tab:foundation_model}, our method consistently improves performance on OpenCLIP across four benchmarks.
Specifically, it achieves an average accuracy gain of \textbf{8.79}~pp over the best-performing baseline for OpenCLIP across the three imbalance ratios.
These results indicate that the proposed model serves as a generalized solution that enhances the robustness of various VLMs against long-tailed distributions.
Results with imbalance ratios of 10 and 50 and additional results on SigLIP can be found in Appendix~\ref{app:more_results_foundation}.

\begin{table}[!t]
\captionsetup{skip=0pt}
\caption{Ablation results of Boundary-Preserving Alignment (BPA) and Margin-aware Self-Refinement (MSR) with different imbalance ratios. The improvement is indicated by $\uparrow$.}
\vspace{0.15cm}
\label{tab:ablation}
\centering

\resizebox{0.9\linewidth}{!}{
    \setlength{\tabcolsep}{6pt}
    \begin{tabular}{cc|ccc|ccc|ccc|l}
    \toprule

    \multirow{2}{*}[-3pt]{BPA}
    & \multirow{2}{*}[-3pt]{MSR}
    & \multicolumn{3}{c|}{\textbf{Imbalance Ratio = 10}}
    & \multicolumn{3}{c|}{\textbf{Imbalance Ratio = 20}}
    & \multicolumn{3}{c|}{\textbf{Imbalance Ratio = 50}}
    & \multicolumn{1}{c}{\multirow{2}{*}[-3pt]{Avg.}} \\

    \cmidrule(lr){3-5}
    \cmidrule(lr){6-8}
    \cmidrule(lr){9-11}

    & & Res. & Flw. & ES.
      & Res. & Flw. & ES.
      & Res. & Flw. & ES.
      & \\
    \midrule

    \xmark & \xmark
    & 54.48 & 63.67 & 32.88
    & 54.48 & 63.67 & 32.88
    & 54.48 & 63.67 & 32.88
    & 50.34 \\

    \cmark & \xmark
    & 67.05 & 65.23 & 57.26
    & 64.69 & 63.98 & 61.92
    & 63.34 & 64.20 & 50.68
    & \imp{62.04}{11.70} \\

    \xmark & \cmark
    & 61.50 & 68.42 & 52.16
    & 61.71 & 67.33 & 52.26
    & 58.65 & 66.92 & 52.18
    & \imp{60.13}{\phantom{0}9.79} \\

    \cmark & \cmark
    & \textbf{69.43} & \textbf{69.41} & \textbf{74.40}
    & \textbf{68.52} & \textbf{67.73} & \textbf{74.22}
    & \textbf{68.66} & \textbf{67.98} & \textbf{73.68}
    & \textbf{70.45}\hspace{5pt}$\uparrow$\hspace{1pt}{\scriptsize 20.11} \\

    \bottomrule
    \end{tabular}
}
\vspace{-0.1cm}
\end{table}

\subsection{Ablation Studies}

\textbf{All components contribute to the proposed method.}
As shown in Table~\ref{tab:ablation}, both Boundary-Preserving Alignment (BPA) and Margin-aware Self-Refinement (MSR) individually result in noticeable performance improvements over zero-shot CLIP across all settings.
Specifically, BPA and MSR yield substantial average improvements of \textbf{11.70}~pp and \textbf{9.79}~pp, respectively.
Compared with MSR, BPA exhibits a larger improvement.
A possible reason is that BPA mitigates feature drift induced by noisy pseudo-labels and effectively prevents severe boundary erosion, which is the primary cause of performance degradation.
Moreover, the combination of the two components consistently yields the highest accuracy, with a total boost of \textbf{20.11}~pp, highlighting their complementary effects on improving feature discriminability.
The full model also stays stable across imbalance ratios, dropping only \textbf{0.97}~pp from 10 to 50.
We further examine the impact of visual encoder capacity with a ViT-B/16 backbone in Appendix~\ref{app:more_results_vit-b-16}, where MARS outperforms the strongest baseline by \textbf{3.82}~pp on average.

\subsection{More Analyses}
\label{sec:more_analyses}

\textbf{Does boundary-preserving alignment (BPA) effectively reduce head-class boundary erosion?}
To quantify the impact of BPA, we visualize the density curve of logit margin changes ($\Delta m$) for CPL
\setlength{\columnsep}{0.3cm}
\begin{wrapfigure}[13]{r}{0.48\linewidth}
    \centering
    \renewcommand{\thesubfigure}{\alph{subfigure}}
    \captionsetup[subfigure]{labelfont={normalfont}, textfont={normalfont}, skip=0pt}
    \vspace{-0.2cm}

    \includegraphics[width=\linewidth]{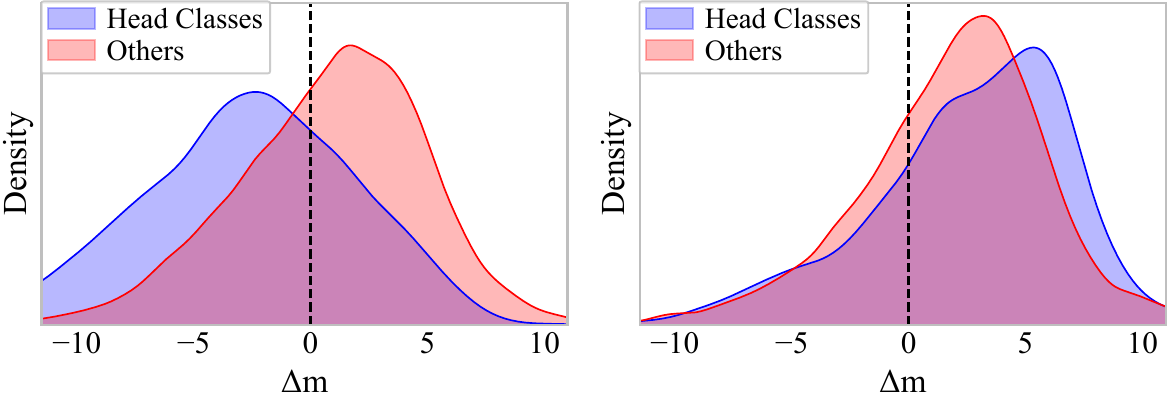}

    \vspace{-0.1cm}

    \begin{subfigure}[t]{0.495\linewidth}
        \caption{Existing method}
        \label{fig:5_a}
    \end{subfigure}%
    \hfill
    \begin{subfigure}[t]{0.495\linewidth}
        \caption{Ours}
        \label{fig:5_b}
    \end{subfigure}

    \vspace{-0.15cm}

    \caption{Density curve of logit margin changes ($\Delta m$). (a) Existing methods cause a negative shift for head classes. (b) Our method maintains positive margin changes.}
    \label{fig:5}

    \vspace{-0.3cm}
\end{wrapfigure}
and our method in Fig.~\ref{fig:5}.
The negative shift in the head-class margin reveals that CPL enforces a uniform constraint at the cost of eroding the head-class boundary.
In contrast, our method rectifies this issue by ensuring that the margin distributions for all classes are shifted to the positive region ($\Delta m > 0$), preventing boundary erosion. BPA also reduces head-to-tail misassignment: on RESISC45 with $\gamma=50$, CPL increases the number of head-class samples predicted as tail classes from 428 to 1,060, whereas BPA reduces it to 392. More datasets and a sample-level margin analysis can be found in Appendix~\ref{app:h2t}.

\begin{wrapfigure}[13]{r}{0.48\linewidth}
    \centering
    \renewcommand{\thesubfigure}{\alph{subfigure}}
    \captionsetup[subfigure]{labelfont={normalfont}, textfont={normalfont}, skip=0pt}
    \vspace{-0.25cm}

    \includegraphics[width=\linewidth]{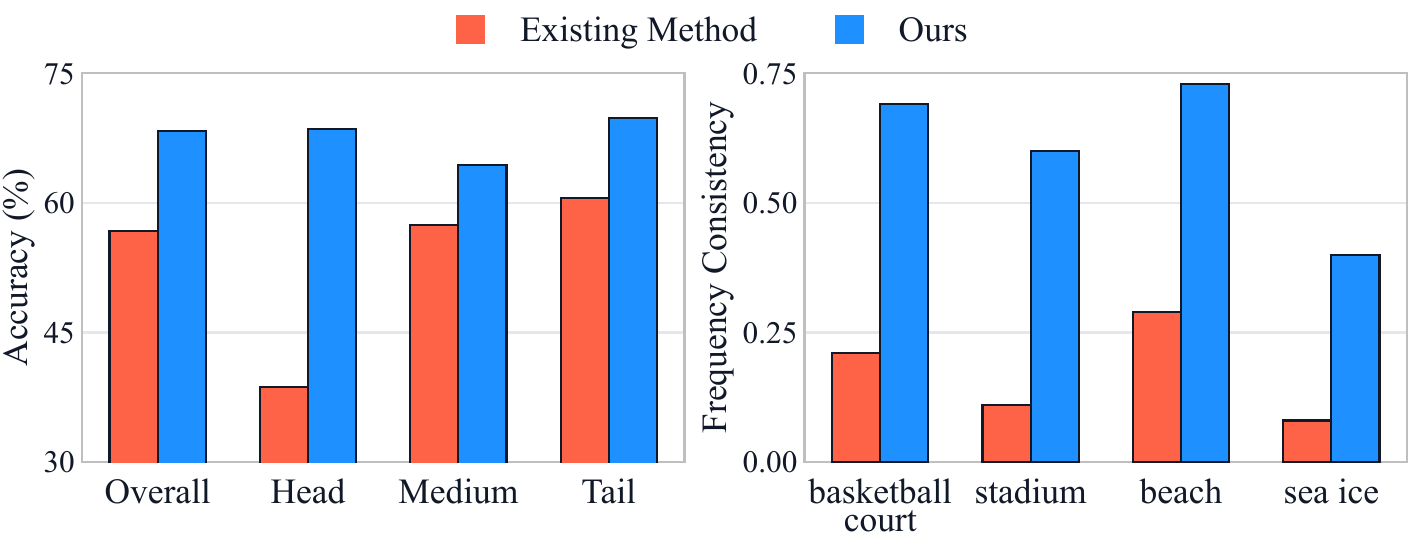}

    \vspace{-0.15cm}

    \begin{subfigure}[t]{0.495\linewidth}
        \caption{}
        \label{fig:6_a}
    \end{subfigure}%
    \hfill
    \begin{subfigure}[t]{0.495\linewidth}
        \caption{}
        \label{fig:6_b}
    \end{subfigure}

    \vspace{-0.15cm}

    \caption{Evaluation of margin-aware self-refinement on RESISC45. (a) Accuracy across head, medium, and tail groups. (b) Frequency consistency on confusable classes.}
    \label{fig:6}

    \vspace{-0.3cm}
\end{wrapfigure}

\textbf{Does margin-aware self-refinement (MSR) effectively improve tail and confusable classes without compromising head classes?}
As shown in Fig.~\ref{fig:6}, to verify whether MSR effectively addresses
\setlength{\columnsep}{0.3cm}
this dilemma, we analyze the performance across different class groups and the frequency consistency (i.e., the alignment between predicted and ground-truth class distributions) in confusable classes.
Specifically, our method achieves substantial accuracy gains in the tail and medium groups while preserving performance on head classes, confirming that discriminability is enhanced without knowledge forgetting.
Furthermore, the significantly higher frequency consistency in confusable classes demonstrates that MSR effectively rectifies prediction errors caused by the model's inherent bias.
More fine-grained results and analyses can be found in Appendix~\ref{app:fine_grained_results}.

\subsection{More Challenging Settings}

\textbf{Our method remains effective on large-scale long-tailed benchmarks.}
To further evaluate the
\setlength{\columnsep}{0.25cm}
\begin{wraptable}{r}{0.48\linewidth}
\centering
\vspace{-0.3cm}

\captionsetup{skip=2pt}
\caption{Comparison of test accuracy (\%) on large-scale long-tailed datasets.}
\label{tab:large_scale_lt}

\vspace{0.05cm}

\resizebox{\linewidth}{!}{
\setlength{\tabcolsep}{6pt}
\begin{tabular}{l|ccc|c}
\toprule
Method
& Places-LT
& ImageNet-LT
& iNaturalist2018
& Avg. \\
\midrule
ZS CLIP
& 38.33 & 59.71 & 69.33 & 55.79 \\
\midrule
FPL
& 39.31 & 60.05 & 69.50 & 56.29 \\
GRIP
& 36.64 & 59.92 & 66.60 & 54.39 \\
CPL
& 38.38 & 55.77 & 65.30 & 53.15 \\
UEO
& 39.71 & 60.23 & 70.17 & 56.70 \\
TMP
& 32.18 & 48.32 & 42.17 & 40.89 \\
CAP
& 37.86 & 61.34 & 70.00 & 56.40 \\
microCLIP
& 40.39 & 58.62 & 69.00 & 56.00 \\
\midrule
\rowcolor[HTML]{E8E5E5}
\algo
& \textbf{41.35} & \textbf{62.96} & \textbf{73.33} & \textbf{59.21} \\
\bottomrule
\end{tabular}
}

\vspace{-0.35cm}
\end{wraptable}
scalability and robustness of our method, we run additional experiments on large-scale long-tailed benchmarks~\citep{liu2019largelt, yang2022ltsurvey, van2018inaturalist, van2021inaturalist} (Places-LT, ImageNet-LT, and an iNaturalist2018 subset, with imbalance ratios of 996, 256, and 500).
As shown in Table~\ref{tab:large_scale_lt}, MARS achieves the best performance on all three benchmarks, improving the average accuracy by \textbf{2.51}~pp.
Notably, it improves the best baseline by \textbf{3.16}~pp on iNaturalist2018.
Base-to-new generalization results are provided in Appendix~\ref{app:base_to_new}.

\section{Conclusion}
\label{sec:conclusion}

In this paper, we investigate unsupervised long-tailed adaptation (ULTA) of vision-language models.
We find that imposing uniform constraints erodes the head-class boundary while reinforcing prediction bias in confusable classes.
To address this, we propose margin-aware refinement with structural alignment (MARS), which integrates boundary-preserving alignment to maintain the head-class boundary with margin-aware self-refinement to enhance the discriminability of tail and confusable classes.
Extensive experiments across nine benchmarks with varying imbalance ratios validate the effectiveness of MARS, with an average improvement of \textbf{4.71} percentage points.

\bibliography{references}
\bibliographystyle{iclr2027_conference}

\newpage
\appendix

\section*{Table of Contents for the Appendix}
\begin{itemize}[leftmargin=0pt, label={}, itemsep=8pt]
    \item \ref{app:training_process}\quad Training Process of MARS \dotfill \pageref{app:training_process}
    \item \ref{app:datasets_details}\quad Dataset Details \dotfill \pageref{app:datasets_details}
    \item \ref{app:implementation_details}\quad Implementation Details \dotfill \pageref{app:implementation_details}
    \item \ref{app:more_experiment}\quad More Experiment Results and Analyses \dotfill \pageref{app:more_experiment}
    \begin{itemize}[leftmargin=20pt, label={}, itemsep=4pt, topsep=6pt]
        \item \ref{app:sensitivity_hyperparameters}\quad Sensitivity of Hyperparameters \dotfill \pageref{app:sensitivity_hyperparameters}
        \item \ref{app:statistical_significance}\quad Statistical Significance Analysis \dotfill \pageref{app:statistical_significance}
        \item \ref{app:more_results_foundation}\quad Comparison with Different VLMs \dotfill \pageref{app:more_results_foundation}
        \item \ref{app:more_results_vit-b-16}\quad Comparison with Different Visual Backbones \dotfill \pageref{app:more_results_vit-b-16}
        \item \ref{app:fine_grained_results}\quad Fine-grained Results across Head, Medium, and Tail Groups \dotfill \pageref{app:fine_grained_results}
        \item \ref{app:h2t}\quad Head-to-Tail Misassignment Analysis \dotfill \pageref{app:h2t}
        \item \ref{app:base_to_new}\quad Base-to-new Generalization \dotfill \pageref{app:base_to_new}
    \end{itemize}
\end{itemize}

\newpage

\FloatBarrier
\section{Training Process of MARS}
\label{app:training_process}

Algorithm~\ref{alg:simplified_algorithm} provides the complete training procedure of MARS.

\begin{algorithm}[!ht]
\caption{Training Process of MARS}
\label{alg:simplified_algorithm}
\begin{algorithmic}[1]
\STATE \textbf{Input}: Unlabeled dataset $\mathcal{D}_u$, alignment epochs $T_{align}$, and maximum epochs $T_{max}$.
\STATE \textbf{Output}: Trained prompts $\theta_{visual},\theta_{textual}$ and student textual adapter $\phi_S$.

\STATE Initialize prompts $\theta_{textual},\theta_{visual}$ and teacher textual adapter $\phi_T$.
\STATE Cache zero-shot features and predictions on the weak view, and construct class anchors.

\STATE \textit{// Boundary-Preserving Alignment (Sec.~\ref{sec:alignment})}
\FOR{epoch = 1 \textbf{to} $T_{align}$ }

            \STATE For all $x_i\in\mathcal D_u$, obtain frozen $a_i$ and current $p_i$ on the same weak view.
            \STATE Compute visual conditions $h_{ij}$, pairwise coefficients $\rho_{cj}$ by Eq.~\eqref{eq:bpa_direction}, and targets by Eq.~\eqref{eq:bpa_target}.
            \STATE With fixed $q_i$, update $\theta_{textual},\theta_{visual},\phi_T$ on strong-view minibatches via Eq.~\eqref{eq:bpa_loss}.

\ENDFOR

\STATE \textit{// Margin-aware Self-Refinement (Sec.~\ref{sec:refinement})}

\STATE Freeze teacher's predictions, $\theta_{visual}$ and $\phi_T$; compute class prior $r_{\hat{y}}$ and initialize $\phi_S$.

\FOR{epoch = $T_{align}+1$ \textbf{to} $T_{max}$ }
            \STATE Obtain visual feature $\mathbf{v}$ via frozen $\theta_{visual}$ and textual embedding $\mathbf{t}$ via $\theta_{textual}\rightarrow\phi_S$;
            \STATE Update CMC and compute gating factor $\hat{\alpha}$ by Eq.~\eqref{eq:alpha};
            \STATE Compute $\mathcal{L}_{\text{refine}}$ and update $\theta_{textual},\phi_S$ via gradient descent;

\ENDFOR
\end{algorithmic}
\end{algorithm}

\FloatBarrier
\section{Dataset Details}
\label{app:datasets_details}

We evaluate our method on a diverse set of benchmark datasets to demonstrate its robustness across different domains. Table \ref{tab:data_appendix} summarizes the detailed statistics of these datasets, including the class count and the number of samples in training and test splits. Note that we include datasets with both balanced and imbalanced test sets to conduct a comprehensive evaluation.

\begin{table}[!ht]
\caption{Statistics for benchmark datasets used in main experiments. The rightmost column indicates whether the testing set is balanced.}
\label{tab:data_appendix}
\centering
\vspace{0.15cm}

{\small
    \begin{tabular}{l|cccc}
    \toprule
    Dataset & Classes & Train & Test & Balanced \\
    \midrule
    CUB \citep{wah2011cub}       &  200       &  5,994       &  5,794    &  \xmark      \\
    RESISC45 \citep{cheng2017res}        &  45         &  6,300       &  25,200    &  \cmark      \\
    FGVCAircraft \citep{maji2013fa}      &  100        &  6,667       &  3,333    &  \cmark      \\
    Caltech101 \citep{fei2004cal}        &  100        &  5,777       &  2,465    &  \xmark      \\
    EuroSAT \citep{2019es}               &  10         &  22,000      &  5,000    &  \cmark      \\
    Flowers102 \citep{2008flw}           &  102        &  2,040       &  6,149    &  \xmark      \\
    Food101 \citep{bossard2014food}      &  101        &  70,700      &  30,300    &  \cmark      \\
    OxfordPets \citep{2012pets}          &  37         &  3,680       &  3,669    &  \xmark      \\
    StanfordCars \citep{fei2013cars}     &  196        &  8,144       &  8,041    &  \xmark      \\
    \bottomrule
    \end{tabular}
}

\end{table}

\FloatBarrier
\section{Implementation Details}
\label{app:implementation_details}

We rerun all baseline methods from their official public codebases under a unified protocol. Specifically, we standardize the visual backbone to OpenAI CLIP ViT-B/32 and use the same training dataset, data split, and evaluation metrics for all methods.
All other settings follow the recommendations in the corresponding official implementations, such as the training schedule and optimization hyperparameters.
For MARS, in Stage 1, we use a batch size of 64 and train the model for 10 epochs with the SGD optimizer, optimizer momentum 0.9, initial learning rate 0.0035, weight decay 5e-4, and a linear-warmup cosine learning-rate schedule.
In Stage 2, we train the student model for 30 epochs.
We fix $\beta=1$ and the momentum coefficient $\lambda=0.95$ for all datasets.
In the balanced setting of Table~\ref{tab:balanced}, we further set the imbalance ratio to 1 (cap the maximum number of training samples per class at 100).
We conduct the experiments on a single NVIDIA RTX 4090 GPU using PyTorch v2.8.0.

\FloatBarrier
\section{More Experiment Results and Analyses}
\label{app:more_experiment}

\FloatBarrier
\subsection{Sensitivity of Hyperparameters}
\label{app:sensitivity_hyperparameters}

\begin{figure}[!ht]
    \centering
    \includegraphics[width=0.95\linewidth]{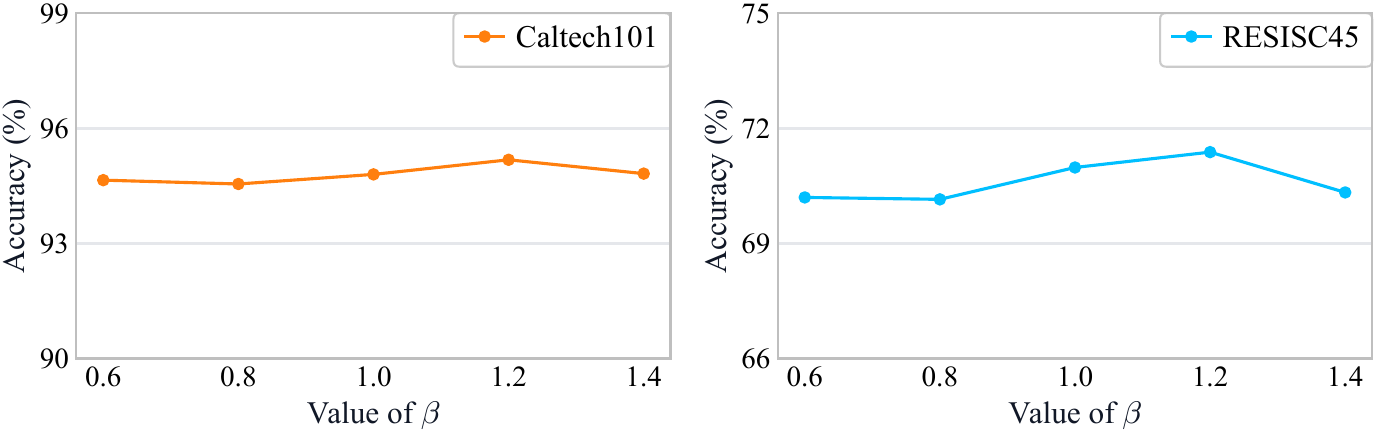}

    \vspace{-0.1cm}

    \caption{Sensitivity to $\beta$ on Caltech101 (left) and RESISC45 (right).}
    \label{fig:7}

    \vspace{-0.1cm}
\end{figure}

In the previous experiments, we fix $\beta=1$ across datasets. We evaluate the sensitivity of this exponent in Fig.~\ref{fig:7} on Caltech101 and RESISC45.

\FloatBarrier
\subsection{Statistical Significance Analysis}
\label{app:statistical_significance}

To further validate the statistical significance of the improvements, we conduct two-sided paired t-tests at a 0.05 significance level against seven baseline methods using the three corresponding seed results. As shown in Table~\ref{tab:significance}, our method achieves significant improvements in \textbf{84.7\%} of cases (160 out of 189), confirming its effectiveness under different imbalance ratios.

\begin{table}[!ht]
    \centering
    \captionsetup{skip=0pt}
    \caption{Statistical significance of performance differences assessed with a two-sided paired t-test at a 0.05 significance level over three corresponding seeds, reported as win/tie/loss counts. Imbalance Ratio is abbreviated as IR.}
    \label{tab:significance}

    \vspace{0.15cm}

    \resizebox{0.5\linewidth}{!}{
        \setlength{\tabcolsep}{5pt}
        \begin{tabular}{l|ccc|c}
        \toprule
        Method & IR = 10 & IR = 20 & IR = 50 & Total \\
        \midrule
        FPL  & 8 / 1 / 0 & 9 / 0 / 0 & 9 / 0 / 0 & 26 / 1 / 0 \\
        GRIP & 9 / 0 / 0 & 8 / 1 / 0 & 8 / 1 / 0 & 25 / 2 / 0 \\
        CPL  & 8 / 1 / 0 & 8 / 1 / 0 & 8 / 1 / 0 & 24 / 3 / 0 \\
        UEO  & 9 / 0 / 0 & 6 / 3 / 0 & 7 / 2 / 0 & 22 / 5 / 0 \\
        TMP  & 9 / 0 / 0 & 9 / 0 / 0 & 9 / 0 / 0 & 27 / 0 / 0 \\
        CAP  & 4 / 5 / 0 & 7 / 2 / 0 & 7 / 2 / 0 & 18 / 9 / 0 \\
        microCLIP & 6 / 3 / 0 & 7 / 2 / 0 & 5 / 4 / 0 & 18 / 9 / 0 \\
        \midrule
        Total & 53 / 10 / 0 & 54 / 9 / 0 & 53 / 10 / 0 & 160 / 29 / 0 \\
        \bottomrule
        \end{tabular}
    }

    \vspace{-0.2cm}
\end{table}

\FloatBarrier
\subsection{Comparison with Different VLMs}
\label{app:more_results_foundation}
To verify the generalization capability of our approach, we complement the OpenCLIP results in Table~\ref{tab:foundation_model} with imbalance ratios of 10 and 50 (Table~\ref{tab:foundation_model_openclip}), and further replace CLIP with SigLIP (Table~\ref{tab:foundation_model_2}). It can be observed that our method maintains a significant performance advantage over previous baselines on both VLMs, demonstrating that our strategy is effective across different VLMs. Notably, with SigLIP, our method surpasses the best-performing baseline by \textbf{5.10}~pp on RESISC45 under the imbalance ratio of 50.

\begin{table}[!ht]
\captionsetup{skip=0pt}
\caption{Comparison of test accuracy (\%) using OpenCLIP with imbalance ratios of 10 and 50. Zero-shot OpenCLIP is abbreviated as ZS OpenCLIP.}
\label{tab:foundation_model_openclip}
\centering
\vspace{0.15cm}

\setlength{\tabcolsep}{9pt}
\resizebox{0.9\linewidth}{!}{
    \begin{tabular}{l|cccc|cccc|c}
    \toprule
    \multirow{2}{*}[-3pt]{Method} & \multicolumn{4}{c|}{\textbf{Imbalance Ratio = 10}} & \multicolumn{4}{c|}{\textbf{Imbalance Ratio = 50}} & \multirow{2}{*}[-3pt]{Avg.} \\
    \cmidrule(lr){2-5} \cmidrule(lr){6-9}
     & Res. & Flw. & ES. & Pets. & Res. & Flw. & ES. & Pets. & \\
    \midrule
    ZS OpenCLIP & 63.90 & 71.43 & 51.74 & 90.62 & 63.90 & 71.43 & 51.74 & 90.62 & 69.42 \\
    \midrule
    FPL & 64.05 & 66.43 & 50.10 & 88.03 & 59.27 & 65.18 & 49.98 & 86.92 & 66.25 \\
    GRIP & 64.63 & 64.29 & 62.08 & 82.47 & 57.94 & 66.12 & 60.20 & 81.22 & 67.37 \\
    CPL & 66.66 & 63.20 & 64.18 & 86.67 & 59.65 & 63.70 & 60.42 & 81.38 & 68.23 \\
    UEO & 62.49 & 70.16 & 53.54 & 88.66 & 58.77 & 70.22 & 51.44 & 89.02 & 68.04 \\
    TMP & 45.98 & 16.34 & 10.08 & 60.70 & 34.51 & 11.79 & 10.00 & 35.43 & 28.10 \\
    CAP & 67.60 & 65.95 & 65.36 & 88.28 & 58.32 & 61.67 & 50.36 & 73.45 & 66.37 \\
    microCLIP & 66.44 & 67.52 & 59.30 & 89.62 & 64.19 & 65.46 & 57.52 & 88.96 & 69.88 \\
    \midrule
    \rowcolor[HTML]{E8E5E5} \algo & \textbf{72.10} & \textbf{75.64} & \textbf{76.74} & \textbf{92.12} & \textbf{70.10} & \textbf{73.52} & \textbf{72.34} & \textbf{91.77} & \textbf{78.04} \\
    \bottomrule
    \end{tabular}
}
\end{table}

\begin{table}[!ht]
\captionsetup{skip=0pt}
\caption{Comparison of test accuracy (\%) using SigLIP with varying imbalance ratios (10, 20, 50). Zero-shot SigLIP is abbreviated as ZS SigLIP.}
\label{tab:foundation_model_2}
\centering
\vspace{0.15cm}

\setlength{\tabcolsep}{5pt}
\resizebox{\linewidth}{!}{
    \begin{tabular}{l|cccc|cccc|cccc|c}
    \toprule
    \multirow{2}{*}[-3pt]{Method} & \multicolumn{4}{c|}{\textbf{Imbalance Ratio = 10}} & \multicolumn{4}{c|}{\textbf{Imbalance Ratio = 20}} & \multicolumn{4}{c|}{\textbf{Imbalance Ratio = 50}} & \multirow{2}{*}[-3pt]{Avg.} \\
    \cmidrule(lr){2-5} \cmidrule(lr){6-9} \cmidrule(lr){10-13}
     & Res. & Flw. & ES. & Pets. & Res. & Flw. & ES. & Pets. & Res. & Flw. & ES. & Pets. & \\
    \midrule
    ZS SigLIP & 59.23 & 83.46 & 35.74 & 93.10 & 59.23 & 83.46 & 35.74 & 93.10 & 59.23 & 83.46 & 35.74 & 93.10 & 67.88 \\
    \midrule
    FPL      & 61.14 & 82.96 & 37.12 & 92.80 & 60.04 & 83.36 & 35.14 & 92.45 & 60.23 & 83.48 & 35.62 & 92.83 & 68.10 \\
    GRIP     & 62.36 & 83.72 & 45.38 & 91.52 & 61.73 & 83.66 & 42.62 & 91.71 & 60.62 & 83.72 & 34.72 & 91.11 & 69.41 \\
    CPL      & 65.56 & 81.33 & 46.74 & 90.92 & 62.76 & 81.22 & 52.14 & 88.91 & 60.50 & 82.50 & 44.24 & 88.72 & 70.46 \\
    UEO      & 59.65 & 83.48 & 40.10 & 93.08 & 59.55 & 83.46 & 39.94 & 93.10 & 59.40 & 83.46 & 39.74 & 93.08 & 69.00 \\
    TMP      & 19.25 & 7.82 & 16.98 & 28.75 & 17.05 & 6.05 & 15.36 & 20.58 & 14.27 & 3.85 & 16.84 & 13.44 & 15.02 \\
    CAP      & 72.62 & 84.48 & 50.18 & 86.83 & 69.36 & 85.15 & 46.86 & 84.73 & 63.03 & 83.59 & 47.64 & 82.03 & 71.38 \\
    microCLIP & 69.64 & 85.41 & 56.02 & 93.49 & 68.00 & 84.45 & 48.94 & 92.61 & 65.59 & 82.45 & 46.62 & 92.31 & 73.79 \\
    \midrule
    \rowcolor[HTML]{E8E5E5} \algo  & \textbf{75.15} & \textbf{89.98} & \textbf{68.36} & \textbf{94.99} & \textbf{74.03} & \textbf{88.96} & \textbf{71.92} & \textbf{95.28} & \textbf{70.69} & \textbf{89.15} & \textbf{61.46} & \textbf{94.88} & \textbf{81.24} \\
    \bottomrule
    \end{tabular}
}
\end{table}

\FloatBarrier
\subsection{Comparison with Different Visual Backbones}
\label{app:more_results_vit-b-16}
We further investigate the scalability of our model by upgrading the visual backbone to a higher-capacity ViT-B/16 encoder. The results presented in Table~\ref{tab:vit_b_16} indicate that MARS maintains a performance advantage over existing baselines under all imbalance ratios. Specifically, our method achieves the highest average accuracy of \textbf{74.41\%}, outperforming the strongest baseline by \textbf{3.82}~pp on average, with consistent gains across all four datasets. This validates that our approach is not only robust across different VLMs but also capable of effectively leveraging the enhanced features provided by the stronger vision backbone.

\begin{table}[!ht]
\captionsetup{skip=0pt}
\caption{Comparison of test accuracy (\%) using ViT-B/16 as the visual backbone with imbalance ratios of 10, 20, and 50.}
\label{tab:vit_b_16}
\centering
\vspace{0.15cm}

\setlength{\tabcolsep}{5pt}
\resizebox{\linewidth}{!}{
    \begin{tabular}{l|cccc|cccc|cccc|c}
    \toprule
    \multirow{2}{*}[-3pt]{Method} & \multicolumn{4}{c|}{\textbf{Imbalance Ratio = 10}} & \multicolumn{4}{c|}{\textbf{Imbalance Ratio = 20}} & \multicolumn{4}{c|}{\textbf{Imbalance Ratio = 50}} & \multirow{2}{*}[-3pt]{Avg.} \\
    \cmidrule(lr){2-5} \cmidrule(lr){6-9} \cmidrule(lr){10-13}
     & Res. & Cub. & Flw. & Pets. & Res. & Cub. & Flw. & Pets. & Res. & Cub. & Flw. & Pets. & \\
    \midrule
    ZS CLIP & 62.25 & 55.32 & 71.15 & 88.20 & 62.25 & 55.32 & 71.15 & 88.20 & 62.25 & 55.32 & 71.15 & 88.20 & 69.23 \\
    \midrule
    FPL & 65.02 & 56.81 & 66.74 & 89.40 & 63.02 & 55.50 & 65.08 & 89.29 & 61.38 & 55.35 & 65.88 & 88.85 & 68.53 \\
    GRIP & 65.47 & 54.17 & 69.18 & 86.26 & 61.22 & 53.86 & 69.52 & 85.88 & 58.41 & 52.12 & 67.67 & 84.08 & 67.32 \\
    CPL & 66.72 & 54.98 & 67.23 & 88.80 & 62.83 & 53.27 & 66.08 & 86.24 & 58.16 & 53.33 & 67.18 & 84.06 & 67.41 \\
    UEO & 61.62 & 57.18 & 70.92 & 89.62 & 60.75 & 56.59 & 70.73 & 89.59 & 60.15 & 56.61 & 70.60 & 89.45 & 69.48 \\
    TMP & 46.50 & 20.04 & 13.21 & 43.66 & 38.52 & 16.07 & 9.73 & 29.93 & 31.21 & 12.72 & 7.25 & 18.86 & 23.98 \\
    CAP & 69.87 & 56.29 & 69.34 & 89.18 & 65.60 & 56.61 & 69.18 & 84.71 & 63.89 & 54.88 & 67.83 & 83.37 & 69.23 \\
    microCLIP & 69.01 & 53.78 & 72.48 & 90.54 & 67.75 & 51.66 & 71.10 & 90.30 & 66.87 & 51.19 & 71.56 & 90.84 & 70.59 \\
    \midrule
    \rowcolor[HTML]{E8E5E5} \algo & \textbf{72.32} & \textbf{59.84} & \textbf{74.87} & \textbf{93.65} & \textbf{70.51} & \textbf{58.56} & \textbf{75.77} & \textbf{93.24} & \textbf{69.26} & \textbf{58.28} & \textbf{74.14} & \textbf{92.45} & \textbf{74.41} \\
    \bottomrule
    \end{tabular}
}
\end{table}

\FloatBarrier
\subsection{Fine-grained Results across Head, Medium, and Tail Groups}
\label{app:fine_grained_results}

To complement the group-level analysis, we report detailed head, medium, and tail results on RESISC45 and EuroSAT in Table~\ref{tab:fine_grained_head_medium_tail}.
The table shows that our method improves not only overall accuracy but also strongly restores head-class performance while maintaining medium- and tail-class performance.
For example, on RESISC45, the head accuracy improves by \textbf{18.65}~pp over UEO and \textbf{32.57}~pp over CAP.

\begin{table}[!ht]
\captionsetup{skip=0pt}
\caption{Comparison of test accuracy (\%) across head, medium, and tail groups on RESISC45 and EuroSAT with an imbalance ratio of 50.}
\label{tab:fine_grained_head_medium_tail}
\centering
\vspace{0.15cm}

\setlength{\tabcolsep}{8pt}
\resizebox{0.85\linewidth}{!}{
    \begin{tabular}{l|cccc|cccc}
    \toprule
    \multirow{2}{*}[-3pt]{Method}
    & \multicolumn{4}{c|}{Res.}
    & \multicolumn{4}{c}{ES.} \\
    \cmidrule(lr){2-5} \cmidrule(lr){6-9}
    & Overall & Head & Medium & Tail
    & Overall & Head & Medium & Tail \\
    \midrule
    FPL
    & 57.61 & 46.59 & 58.52 & 59.72
    & 55.51 & 23.40 & 60.20 & 58.52 \\

    CPL
    & 56.73 & 38.69 & 57.40 & 60.57
    & 58.82 & 13.00 & 65.53 & 63.10 \\

    UEO
    & 56.97 & 51.90 & 54.43 & 59.83
    & 42.42 & 6.80 & 44.67 & 47.23 \\

    CAP
    & 58.78 & 37.98 & 59.88 & 63.05
    & 50.38 & 26.40 & 60.47 & 49.33 \\

    \midrule
    \rowcolor[HTML]{E8E5E5} \algo

    & \textbf{68.70} & \textbf{70.55} & \textbf{67.01} & \textbf{69.11} & \textbf{75.68} & \textbf{78.60} & \textbf{76.60} & \textbf{74.73} \\
    \bottomrule
    \end{tabular}
}
\vspace{-0.2cm}
\end{table}

\FloatBarrier
\subsection{Head-to-Tail Misassignment Analysis}
\label{app:h2t}

To examine the head-class degradation in Section~\ref{sec:misalignment}, we track how true head-class test samples are predicted after adaptation. H-to-H and H-to-T denote the numbers of these samples misclassified as head and tail classes, respectively. As shown in Table~\ref{tab:h2t_flow}, CPL reduces the head accuracy on RESISC45 from 69.20\% to 38.69\% and increases the H-to-T count from 428 to 1,060, while H-to-H decreases from 429 to 164. BPA restores the head accuracy to 72.89\% and reduces the H-to-T count to 392 without sacrificing the tail accuracy. OxfordPets shows the same pattern.

\begin{table}[!ht]
\captionsetup{skip=0pt}
\caption{Head/tail accuracy (\%) and prediction flow of true head-class samples on RESISC45 and OxfordPets with an imbalance ratio of 50.}
\label{tab:h2t_flow}
\centering
\vspace{0.15cm}
{\small
\setlength{\tabcolsep}{6pt}
\begin{tabular}{ll|cc|cc}
\toprule
Dataset & Method & Head Acc. & Tail Acc. & H-to-H & H-to-T \\
\midrule
\multirow{3}{*}{RESISC45} & ZS CLIP & 69.20 & 56.84 & 429 & 428 \\
 & CPL & 38.69 & 60.57 & 164 & 1,060 \\
 & BPA & \textbf{72.89} & \textbf{65.37} & 210 & \textbf{392} \\
\midrule
\multirow{3}{*}{OxfordPets} & ZS CLIP & 86.75 & 91.47 & 28 & 18 \\
 & CPL & 80.12 & 92.14 & 17 & 56 \\
 & BPA & \textbf{88.14} & \textbf{92.24} & 30 & \textbf{16} \\
\bottomrule
\end{tabular}
}
\end{table}

We further analyze all true head-class samples that CPL predicts as tail classes on Caltech101 and Food101 with an imbalance ratio of 50. For each sample $x$ with ground-truth label $y$, we compute the logit margin change relative to zero-shot CLIP, $\Delta m=m_{\text{Adapted}}(x,y)-m_{\text{Zero-shot CLIP}}(x,y)$, where $m(x,y)=z_y(x)-\max_{j\ne y}z_j(x)$. As shown in Table~\ref{tab:h2t_margin}, CPL yields strongly negative margin changes on these samples, whereas BPA turns them positive. Moreover, BPA no longer predicts most of these samples as tail classes and recovers the correct label for 94.12\% of them on Caltech101 and 61.11\% on Food101.

\begin{table}[!ht]
\captionsetup{skip=0pt}
\caption{Average logit margin changes relative to zero-shot CLIP and repairs over all true head-class samples predicted as tail classes by CPL on Caltech101 and Food101 with an imbalance ratio of 50. ``No longer tail'' counts the samples that BPA no longer predicts as tail classes, and ``Exact repairs'' counts those that BPA predicts correctly.}
\label{tab:h2t_margin}
\centering
\vspace{0.15cm}
{\small
\setlength{\tabcolsep}{6pt}
\begin{tabular}{l|c|cc|cc}
\toprule
Dataset & Cases & CPL $\Delta m$ & BPA $\Delta m$ & No longer tail & Exact repairs \\
\midrule
Caltech101 & 204 & $-3.88$ & $+2.28$ & 199 (97.55\%) & 192 (94.12\%) \\
Food101 & 1,579 & $-4.41$ & $+1.04$ & 1,204 (76.25\%) & 965 (61.11\%) \\
\bottomrule
\end{tabular}
}
\end{table}

\FloatBarrier
\subsection{Base-to-new Generalization}
\label{app:base_to_new}

Beyond standard closed-set classification, we further evaluate base-to-new generalization. Following the standard 1:1 base/new split, we train only on unlabeled long-tailed base-class samples and evaluate on the base and new test sets using the harmonic mean. As shown in Table~\ref{tab:base_to_new}, our method consistently outperforms previous methods on both datasets. Specifically, on RESISC45, MARS improves the best prior harmonic mean from \textbf{58.92\%} to \textbf{62.16\%}, yielding a gain of \textbf{3.24}~pp. These results validate that our method preserves strong cross-class generalization under long-tailed adaptation.

\begin{table}[!ht]
\centering

\captionsetup{skip=2pt}
\caption{Base-to-new generalization on RESISC45 and OxfordPets (imbalance ratio=50).}
\label{tab:base_to_new}

\vspace{0.05cm}

\resizebox{0.6\linewidth}{!}{
\setlength{\tabcolsep}{6pt}
\begin{tabular}{l|ccc|ccc}
\toprule
\multirow{2}{*}[-3pt]{Method}
& \multicolumn{3}{c|}{Res.}
& \multicolumn{3}{c}{Pets.} \\
\cmidrule(lr){2-4} \cmidrule(lr){5-7}
& Base & New & H
& Base & New & H \\
\midrule
ZS CLIP
& 55.36 & 53.56 & 54.45
& 84.77 & 83.95 & 84.36 \\

\midrule
FPL
& 69.05 & 51.39 & 58.92
& 83.68 & 86.63 & 85.13 \\

GRIP
& 64.15 & 42.65 & 51.23
& 77.67 & 85.23 & 81.28 \\

CPL
& 66.33 & 45.27 & 53.81
& 82.46 & 85.68 & 84.04 \\

UEO
& 60.02 & 49.66 & 54.35
& 85.06 & 88.59 & 86.79 \\

CAP
& 66.90 & 44.71 & 53.60
& 82.24 & 78.36 & 80.25 \\
microCLIP
& 67.76 & 44.71 & 53.87
& 80.75 & 88.93 & 84.64 \\

\midrule
\rowcolor[HTML]{E8E5E5} \algo

    & \textbf{69.59} & \textbf{56.16} & \textbf{62.16} & \textbf{89.10} & \textbf{89.49} & \textbf{89.29} \\
\bottomrule
\end{tabular}
}

\end{table}

\FloatBarrier
\end{document}